\documentclass[sigconf]{acmart}

\usepackage{multirow}
\usepackage{amsmath}
\DeclareMathOperator*{\argmax}{arg\,max}

\usepackage{caption} 
\AtBeginDocument{%
  \providecommand\BibTeX{{%
    \normalfont B\kern-0.5em{\scshape i\kern-0.25em b}\kern-0.8em\TeX}}}

\copyrightyear{2020}
\acmYear{2020}
\setcopyright{rightsretained}
\acmConference[MM '20]{Proceedings of the 28th ACM International Conference on Multimedia}{October 12--16, 2020}{Seattle, WA, USA}
\acmDOI{10.1145/3394171.3413856}
\acmISBN{978-1-4503-7988-5/20/10}
\begin{document}
\fancyhead{}

%%
%% The "title" command has an optional parameter,
%% allowing the author to define a "short title" to be used in page headers.
\title{Zero-Shot Multi-View Indoor Localization via Graph Location Networks}

\author{Meng-Jiun Chiou}
\affiliation{%
    \institution{National University of Singapore}
    \city{Singapore}
    \country{Singapore}
}
\email{mengjiun.chiou@u.nus.edu}

\author{Zhenguang Liu}
\affiliation{
    \institution{Zhejiang Gongshang University}
    \city{Hangzhou}
    \country{China}
}
\email{liuzhenguang2008@gmail.com}
\authornote{Corresponding author}

\author{Yifang Yin}
\affiliation{
    \institution{National University of Singapore}
    \city{Singapore}
    \country{Singapore}
}
\email{idsyin@nus.edu.sg}

\author{An-An Liu}
\affiliation{
    \institution{Tianjin University}
    \city{Tianjin}
    \country{China}
}
\email{anan0422@gmail.com}

\author{Roger Zimmermann}
\affiliation{
    \institution{National University of Singapore}
    \city{Singapore}
    \country{Singapore}
}
\email{rogerz@comp.nus.edu.sg}

%%
%% By default, the full list of authors will be used in the page
%% headers. Often, this list is too long, and will overlap
%% other information printed in the page headers. This command allows
%% the author to define a more concise list
%% of authors' names for this purpose.
% \renewcommand{\shortauthors}{Chiou et al.}

%%
%% The abstract is a short summary of the work to be presented in the
%% article.
\begin{abstract}
   Indoor localization is a fundamental problem in location-based applications. Current approaches to this problem typically rely on Radio Frequency technology, which requires not only supporting infrastructures but human efforts to measure and calibrate the signal. Moreover, data collection for all locations is indispensable in existing methods, which in turn hinders their large-scale deployment. In this paper, we propose a novel neural network based architecture Graph Location Networks (GLN) to perform infrastructure-free, multi-view image based indoor localization. GLN makes location predictions based on robust location representations extracted from images through message-passing networks. Furthermore, we introduce a novel zero-shot indoor localization setting and tackle it by extending the proposed GLN to a dedicated zero-shot version, which exploits a novel mechanism Map2Vec to train location-aware embeddings and make predictions on novel unseen locations. Our extensive experiments show that the proposed approach outperforms state-of-the-art methods in the standard setting, and achieves promising accuracy even in the zero-shot setting where data for half of the locations are not available. The source code and datasets are publicly available.\footnote{\url{https://github.com/coldmanck/zero-shot-indoor-localization-release}}.
\end{abstract}

%%
%% The code below is generated by the tool at http://dl.acm.org/ccs.cfm.
%% Please copy and paste the code instead of the example below.
%%


%%
%% Keywords. The author(s) should pick words that accurately describe
%% the work being presented. Separate the keywords with commas.

%% A "teaser" image appears between the author and affiliation
%% information and the body of the document, and typically spans the
%% page.
% \begin{teaserfigure}
%   \includegraphics[width=\textwidth]{sampleteaser}
%   \caption{Seattle Mariners at Spring Training, 2010.}
%   \Description{Enjoying the baseball game from the third-base
%   seats. Ichiro Suzuki preparing to bat.}
%   \label{fig:teaser}
% \end{teaserfigure}

%%
%% This command processes the author and affiliation and title
%% information and builds the first part of the formatted document.
\maketitle

\section{Introduction}

\begin{figure}[t!]
\centering
\includegraphics[width=1.0\columnwidth]{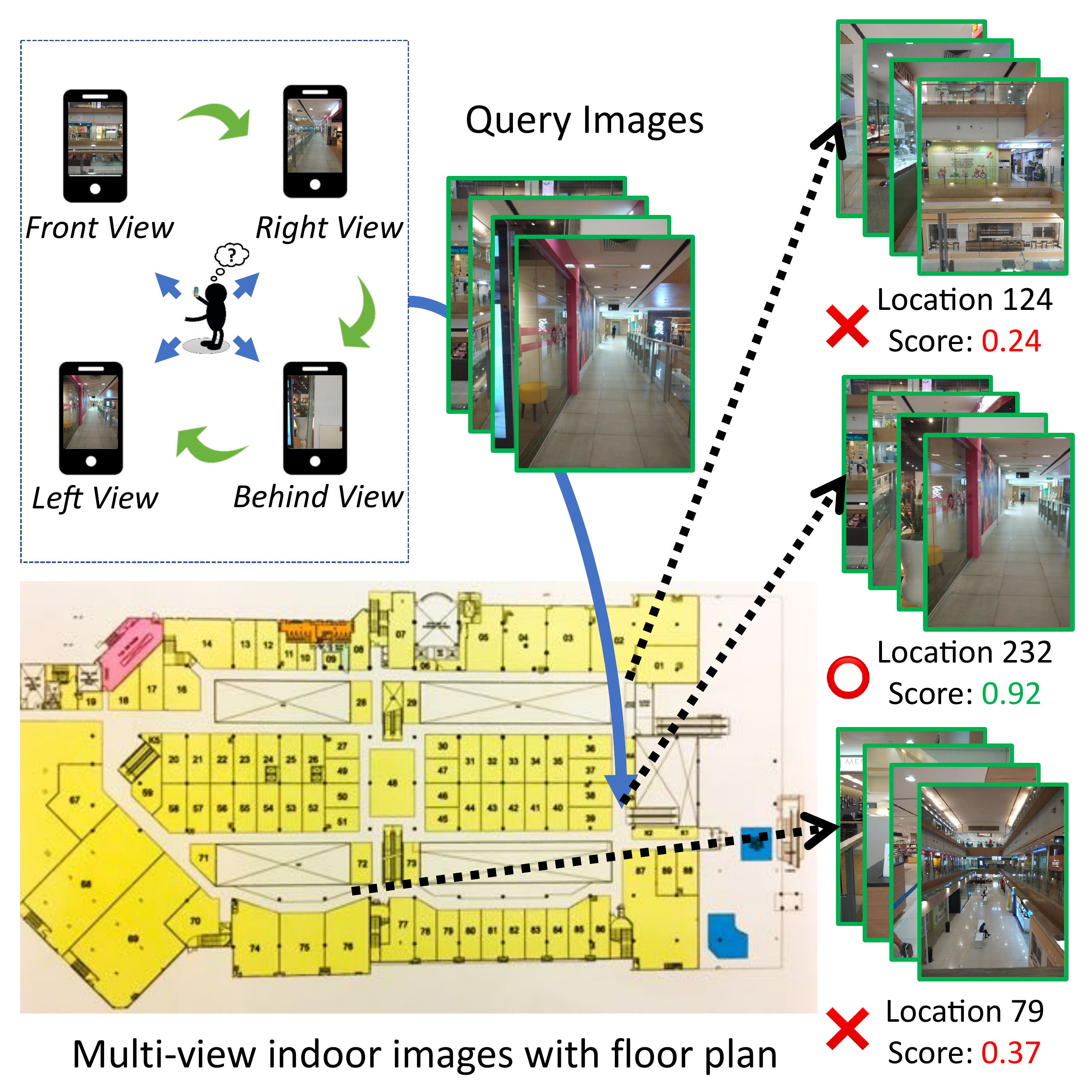} % Reduce the figure size so that it is slightly narrower than the column. Don't use precise values for figure width.This setup will avoid overfull boxes. 
\caption{An illustration of a multi-view image-based indoor localization system. Current location of a user is predicted with the query images (photos of the user's surroundings).}
\label{fig:indoor_localization}
\end{figure}

Indoor localization seeks to localize a user or a device in an indoor environment. Accurate indoor localization systems could enable various applications, \emph{e.g.}, guiding users in an underground parking lot to find a space, and in a large airport to get to the right boarding gate on time \cite{Snoonian2001}; however, it remains to be an open challenge. While Global Positioning System (GPS) has been widely adopted to precisely localize a device in an outdoor environment with a 1- to 5-meter localization accuracy \cite{Karnik2016}, it cannot be simply applied indoors since the GPS signal is significantly weakened after passing through roofs and walls. Researchers have explored other techniques such as WiFi \cite{vasisht2016decimeter,shu2015magicol}, radio frequency identification (RFID) \cite{holm2009hybrid}, optics \cite{liu2010improved}, acoustics \cite{kim2008advanced} and magnetism \cite{shu2015magicol}. However, most of the existing approaches require additional infrastructures, such as WiFi access points, RF transmitters, or specially-designed optic/acoustic receivers. Moreover, manual periodic re-calibration is indispensable for RF-based methods since the signals are prone to fluctuation, which in turn is harder to maintain.

Purely image-based approaches \cite{Liang2013,Gao2016,Ravi2007} are proposed to alleviate part of the deployment costs by utilizing only indoor images. However, most of the existing methods still require special devices to collect data \cite{Liang2013} and utilize fully supervised models to make predictions on each location \cite{Gao2016,Ravi2007}. These limitations cause their approaches to be not only time-consuming but also labor-intensive when deployed in large-scale indoor environments. In this context, an interesting and fundamental question arises: \textit{is it possible to infer the location of the user while data is collected for only several locations?} To answer this question, we will consider \textit{zero-shot learning} where models recognize novel locations by transferring knowledge learned from seen to unseen classes.

Multi-view images are photos of different views at indoor locations and comprise rich location contexts. To leverage this information, a multi-view image- and geomagnetism-based localization strategy is proposed in \cite{liu2017multiview} to transform the problem of indoor localization into a graph retrieval problem. However, they treat different camera views as of equal importance, which is usually not true especially when some views are similar and others are more representative. For example, for two neighboring locations (within one meter) at the same corridor, the views that parallel the corridor are extremely similar while the perpendicular views may consist of more identifiable objects (\emph{e.g.}, different doors and windows). Treating them equally undermines the representativeness of their graph features during retrieval. Moreover, the geomagnetic signals which they rely on are unstable, hard to collect and prone to change with time.

Recently, deep learning \cite{lecun2015deep} has achieved remarkable success in various areas, including but not limited to image-level understanding \cite{simonyan2014very,he2016deep}, object-level detection \cite{ren2015faster,redmon2016you}, human-level estimation \cite{toshev2014deeppose,ruan2019poinet}, text classification \cite{bahdanau2014neural,joulin2016bag} and audio understanding \cite{oord2016wavenet,yin2019multi}. To overcome the aforementioned problems, we exploit the strong representation power of neural networks and propose an infrastructure free, neural network-based architecture Graph Location Networks (GLN) to perform multi-view indoor localization. Given photos of different views, GLN computes robust node representations by aggregating and updating features from neighboring nodes, in which identifiable features permeate the whole graph. A location prediction is made by feeding the representations to a single fully-connected layer. Our proposed approach requires neither any infrastructure nor special devices but only a camera phone to collect the photo database. In addition to evaluating on the publicly available multi-view indoor dataset (\emph{i.e.}, ICUBE \cite{liu2017multiview}), we provide a benchmark dataset WCP that has been collected in a shopping center. We show that our approach outperforms the baseline and existing methods in terms of localization accuracy by a large margin. 

Furthermore, to motivate researches in reducing data collection labor costs we introduce a novel task named \textit{zero-shot indoor localization}, in which half of the locations are masked during training while a system is required to predict the precise user location. We propose a three-step framework to tackle this task and demonstrate the efficiency of it by extending our GLN to a dedicated zero-shot version. Specifically, to transfer the knowledge from the seen locations to the unseen ones, we propose the Map2Vec mechanism that trains location-aware embeddings for both seen and unseen classes by incorporating their geometric contexts of the floor plan. These embeddings are then leveraged to train a compatibility function that maps image-class pairs to scalar scores. Finally, a prediction is made by picking out the best class maximizing the score function of the query image. We demonstrate that, trained through the proposed framework, our model not only surpasses the baseline by a large margin but also achieves promising localization accuracy, \emph{e.g.}, 56.3\% 5-meter accuracy on the ICUBE dataset, while the query locations are never seen during training. To the best of our knowledge, our work is the first exploration of enabling zero shot recognition for indoor localization.

The key contributions of our work are summarized as follows: (a) We propose a novel, neural network based architecture Graph Location Networks which performs effective, infrastructure-free multi-view indoor localization. (b) We introduce zero-shot indoor localization and propose a training framework to tackle it. We demonstrate the efficiency by extending our proposed architecture to a dedicated zero-shot version. (c) We contribute an additional multi-view image-based indoor localization dataset. Our extensive experiments shows that the proposed approach significantly outperforms state-of-the-art methods in the fully supervised setting and achieves competitive localization accuracy in the zero-shot setting.

%%%%%%%%%%%%%%%%%%%%%%%%%%% Related Work %%%%%%%%%%%%%%%%%%%%%%%%%%%

\section{Related Work}
\subsection{Indoor Localization}
Indoor localization has been a popular topic ever since the outdoor localization was mostly tackled \cite{Karnik2016}. Most of the previous efforts rely on RF technology which requires additional transmitters/receivers to estimate the location \cite{vasisht2016decimeter,holm2009hybrid,liu2010improved,kim2008advanced,shu2015magicol} or special devices to collect data \cite{chung2011indoor}, causing large-scale deployment to be costly and prohibitive. More recently and related to our work, a multi-view image- and geomagnetism-based method has been proposed to formulate indoor localization into a graph retrieval problem \cite{liu2017multiview}; however, it does not consider the difference between views and thus fails to capture a robust representation. While purely image-based techniques do not require additional facilities, they either need special devices to do data collection in advance \cite{Liang2013} or require a user to take photos with specific reference objects \cite{Gao2016}. Therefore both are not ideal methods to achieve a pervasive indoor positioning system. In addition, for all existing image-based approaches it is inevitable to collect data of all locations of interest, resulting in costly deployment for large-scale indoor environments. In our work, to implement a truly infrastructure-free indoor localization system, we adopt a purely image-based approach which does not require any special device, only a camera phone, to construct an image database. Furthermore, we introduce zero-shot indoor localization to reduce data collection labor costs. Note that while our method is closely related to outdoor place recognition \cite{arandjelovic2016netvlad,lowry2015visual,sunderhauf2015performance} and is possible to incorporate corresponding techniques (\emph{e.g.} NetVLAD layer \cite{arandjelovic2016netvlad}, contrastive loss \cite{bell2015learning} or 2D-3D hybrid method \cite{sarlin2018leveraging}) to improve the performance, we focus on the graph-based location network with zero-shot setting in this work and leave as possible extensions in future work.

\subsection{Graph-based Methods}
\subsubsection{Graph analysis} 
Graph has garnered a lot of attention from researchers due to its nature of being suitable for representing data in various real-life applications \cite{zhou2018graph}, including protein-protein interaction \cite{fout2017protein}, social relationship networks \cite{hamilton2017inductive}, natural science \cite{sanchez2018graph,battaglia2016interaction}, and knowledge graphs \cite{hamaguchi2017knowledge}. The typical problems that graph analysis is dealing with include node classification, link prediction and clustering. Graph Neural Networks (GNNs) \cite{kipf2016semi,velivckovic2017graph,battaglia2018relational} have become the de facto standard for processing graph-based data for their ability to work on large-scale graphs by borrowing the ideas of weight-sharing and local connections from Convolutional Neural Networks (CNNs) \cite{zhou2018graph}. 

\subsubsection{Graph embedding} 
Nodes in a graph can be represented as feature vectors by incorporating the information of the graph topology and initial node feature \cite{goyal2018graph}. In our work, we leverage GNNs in two scenarios: (a) to perform message passing on a locally-connected location graph for a more robust representation, and (b) to train location embeddings to encode position information for all locations to perform zero-shot recognition. 

\subsection{Zero-Shot Learning}
Unlike traditional supervised learning, zero-shot learning aims to recognize the instance classes that have never been seen by the model during training \cite{Xian2018}. There has been an increasing interest in zero-shot learning and its applications \cite{lampert2013attribute,romera2015embarrassingly,wang2019survey,xian2017zero} since it is not unusual that data is only available for some classes. To transfer knowledge to unseen classes, a compatibility function is learned to relate semantic attributes to features \cite{akata2015evaluation,Sumbul2018}. Specifically, in our work, we learn a compatibility function which maps image features to the location embeddings, and the predicted location is chosen as the one maximizing the compatibility score. Following the definition in \cite{Xian2018}, we perform \textit{generalized zero-shot learning} since our search space contains both training and test classes during testing.

%%%%%%%%%%%%%%%%%%%%%%%%%%% Methodology %%%%%%%%%%%%%%%%%%%%%%%%%%%

\section{Methodology}

In this section, we first formulate the indoor localization problem under fully supervised setting, followed by introducing our proposed method, Graph Location Networks (GLN), which serves as the backbone under both settings. We then demonstrate how to extend our approach to a dedicated zero-shot version to perform indoor positioning on locations of unseen classes.

\subsection{Problem Formulation}
We formulate the image-based indoor localization problem as follows. Given $\mathbf{x} \in \mathcal{X}$ that denotes a set of images and $\mathcal{X}$ the space of all sets of images, we are to predict the location $y \in \mathcal{Y}$ for $\mathbf{x}$, where $\mathcal{Y} = \{y_1,...,y_k\}$ is the set of all $k$ locations. The goal is to learn a function $f:\mathcal{X} \rightarrow \mathcal{Y}$ that maps the input $\mathbf{x}$ to the target class $y$. In our settings, $\mathbf{x}$ comprises images of four different directions, \emph{i.e.}, images of the front, behind, right and left at a location.

\subsection{Standard Graph Location Networks}

\begin{figure*}[t!]
\centering
\includegraphics[width=1.0\textwidth]{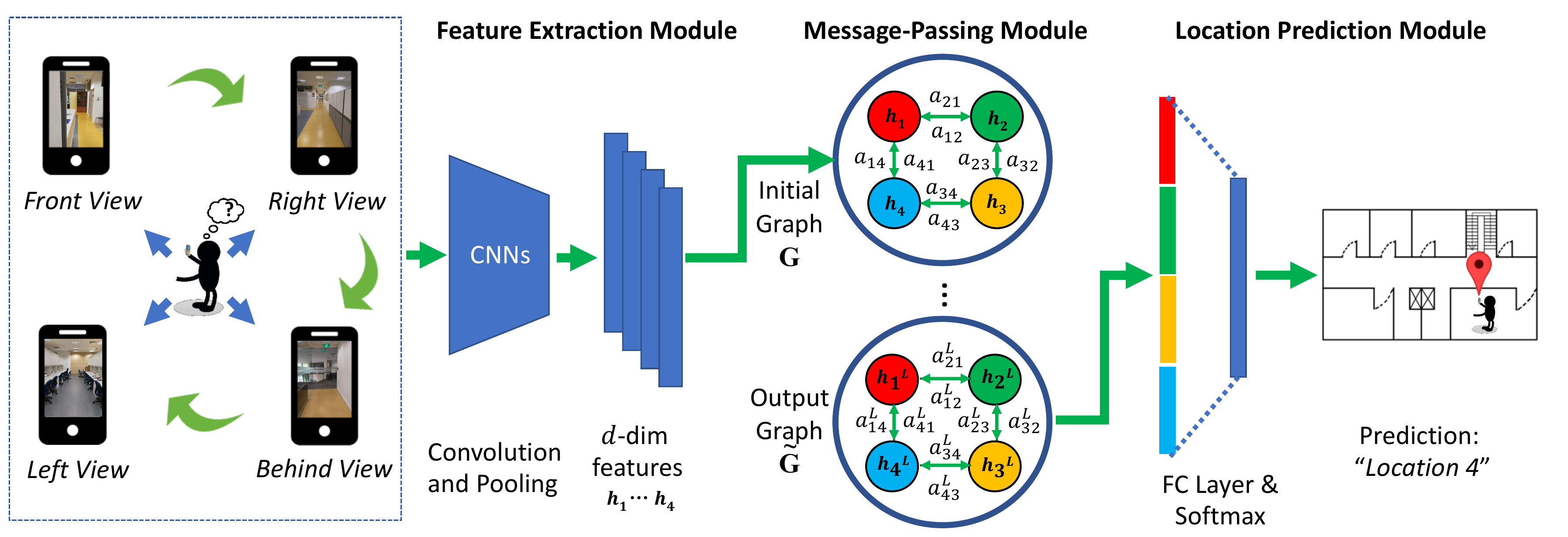}
\caption{The architecture of our Graph Location Networks (GLN) based indoor localization system. First, features of the front, behind, right and left views extracted through CNNs are taken as input by a multi-view quadrilateral graph. An attentional message-passing algorithm is performed on the graph to extract robust location representation, which is then passed into a fully-connected layer followed by a softmax function to make prediction.}
\label{fig:gln}
\end{figure*}

The main idea of our proposed approach is that different views of a location possessing distinct information can be used to form a holistic representation. To take advantage of this, we formulate a locally-connected graph in which features are being refined during the message passing and finally producing a robust location representation for classification. We define Graph Location Networks (GLN) as an indoor localization approach which includes three major modules: feature extraction module, location prediction module, and especially message passing module to exploit the aforementioned graph to make accurate location prediction. We explain each module in detail in the following subsections. Figure \ref{fig:gln} shows an overview of GLN. 

\subsubsection{Feature Extraction Module.}
Given a set of images of the front, behind, right and left views $\mathbf{x} = \{x_1, x_2, x_3, x_4\}$ at a specific location, we utilize a Convolutional Neural Networks based backbone $\varphi$ that takes $\mathbf{x}$ as input to extract high-dimensional features $\mathbf{r} = \varphi(\mathbf{x}) = \{r_1, r_2, r_3, r_4\}, r_i \in \mathcal{R}^d$, where $d$ is the feature dimension. The choice of the backbone network and hyperparameters of the model are given in section \ref{implementation_details}.

\subsubsection{Message Passing Module.}
We define a quadrilateral graph $\mathcal{G}=(\mathcal{V}, \mathcal{E})$ for four views by $\mathcal{V}=\{v_1, v_2, v_3, v_4\}$ and $\mathcal{E}=\{e_{12}, e_{23},$ $e_{34}, e_{41}\}$, where $e_{ij}$ denotes an undirected edge between nodes $v_i$ and $v_j$. Node $v_i$ represents a specific direction and its hidden state is initialized with $r_i$ of that direction. To obtain a robust location representation, our system has to effectively exploit and combine neighboring features. Graph Neural Networks (GNNs) have been shown to be able to aggregate information of neighbor nodes and update the node's hidden state accordingly \cite{scarselli2008graph,kipf2016semi}. We employ GNNs to pass messages within the graph and refine the hidden states of the nodes. Let $h_i^l \in \mathcal{R}^F$ and $h_i^{l+1} \in \mathcal{R}^{F'}$ be the hidden state of node $i$ at layer $l$ and $l+1$, the updating procedure of hidden state $h_i$ of node $v_i$ is defined as follows:
\begin{equation}
    h_i^{l} =
    \begin{cases}
      r_i, & \text{if $l = 1$}\\
      \sigma \left( \sum_{j \in \mathcal{N}(i)} \frac{1}{\alpha_{ij}} W^{l-1} h_j^{l-1} \right), & \text{otherwise}
    \end{cases}
    \label{eq:gcn}
\end{equation}
where $\mathcal{N}(i)$ denotes the set of neighboring nodes of node $v_i$, $\sigma$ is a nonlinear activation function, $\alpha_{ij} = \sqrt{|\mathcal{N}(i)\mathcal{N}(j)|}$ is a normalization constant and $W^l$ represents a shared weight matrix for node-wise feature transformation at layer $l$.

However, each neighboring node (\emph{i.e.}, $\mathcal{N}(i)$) should not have an equal affect to node $i$, \emph{e.g.}, some neighbors may share more overlapped scenes than others. Attention mechanism \cite{vaswani2017attention,velivckovic2017graph} has been demonstrated to be effective to capture relational representation. We introduce a graph self-attention mechanism to assign different weights to each neighbor according to its importance to node $i$. Specifically, we update $h_i$ at layer $l$ with the weight $\alpha_{ij}^l$, which is defined as follows:
\begin{equation}
    \alpha_{ij}^l = \frac{exp(\sigma(a[W^l h_i^l, W^l h_j^l]))}{\sum_{k \in \mathcal{N}(i)} exp(\sigma(a[W^l h_i^l, W^l h_k^l]))},
\end{equation}
where $[\cdot]$ denotes the concatenation operation, and $a$ is a shared attention mechanism that computes the importance of node $j$'s feature to node $i$.

\subsubsection{Location Prediction Module.}
After $L$ layers of message-passing the final hidden states $\mathbf{h}$ are fused to form a single robust representations $\mathbf{x}^r$ as the following:
\begin{equation}
    \mathbf{x}^r = [h_1^L, h_2^L, h_3^L, h_4^L].
\end{equation}
$\mathbf{x}^r$ is then passed into a single fully-connected layer mapping the concatenated feature vector into the location space, followed by a Softmax function to generate a probability distribution over all classes: $p' = f(\text{Softmax}(\text{FC}(\mathbf{x}^r)))$. We adopt the Softmax loss as the objective function as follows:
\begin{equation}
    \begin{aligned}
        \mathcal{L} = -\sum_i p_i\log(p'_i),
    \end{aligned}
\end{equation}
where $p'_i$ is the $i$-th row of $p'$ and $p_i$ denotes the ground truth label for location $i$.

\subsection{Zero-Shot Graph Location Networks}

\begin{figure*}[ht!]
\centering
\includegraphics[width=1.0\textwidth]{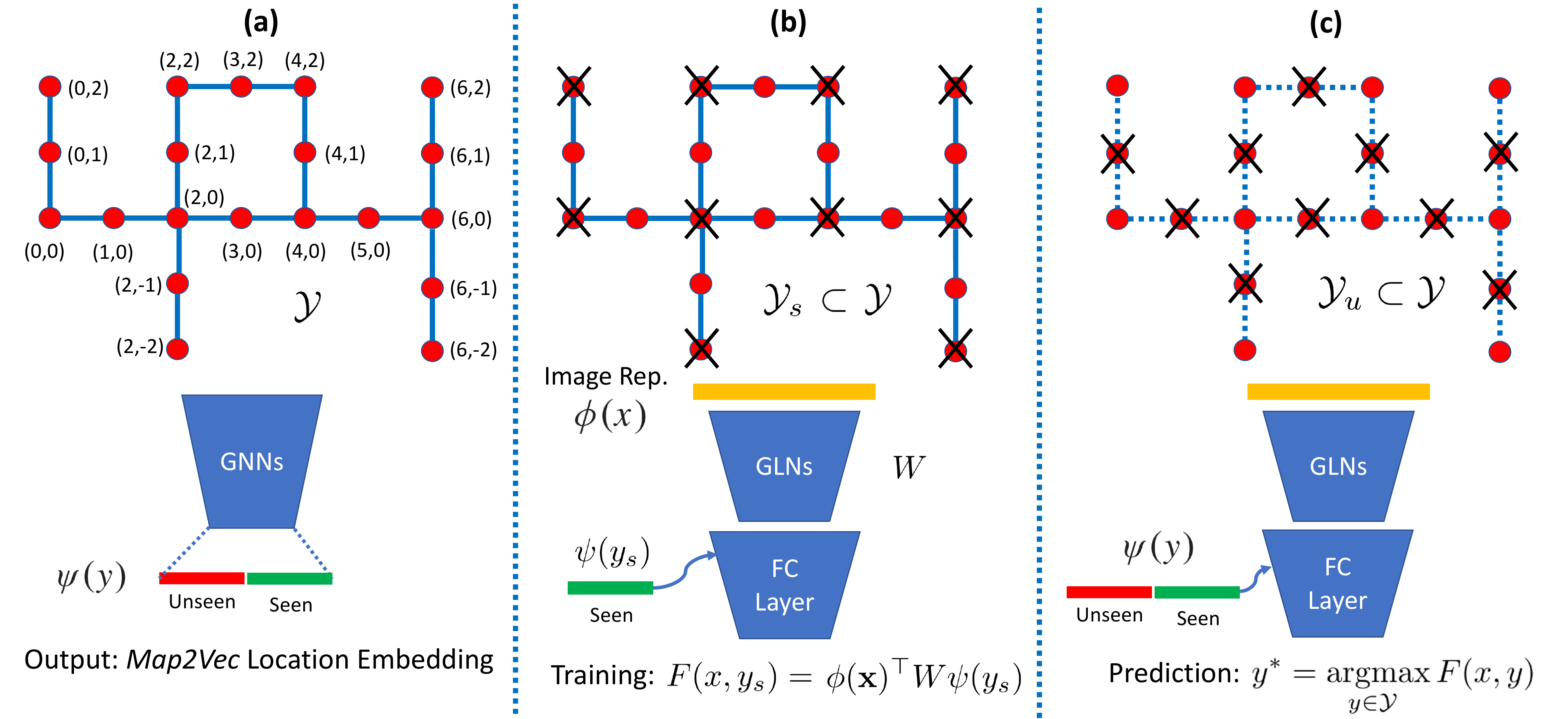}
\caption{The key steps for training a zero-shot indoor localization model. (a) Train Map2Vec location embeddings for a given map (floor plan). (b) Learn a compatibility function with only the seen classes (circles without multiplication sign). (c) Perform zero-shot prediction by assigning an input to the location that maximizes the compatibility function. Dotted lines mean that the edge information is not available during testing. The "GLNs" block can be replaced with any indoor localization model.} 
\label{fig:zero-shot-gln}
\end{figure*}

In this section, we describe the proposed learning framework that enables indoor localization models to perform zero-shot prediction. We use our proposed GLN as the backbone model.

In the zero-shot setting, $\mathcal{Y}$ is divided into two disjointed sets: $\mathcal{Y}_{s} \subset \mathcal{Y}, y_{s} \in \{y_1,...,y_{n}\}$ denotes a set of $n$ \textit{seen classes}, and $\mathcal{Y}_{u} \subset \mathcal{Y}, y_{u} \in \{y_{n+1},...,y_{k}\}$ represents a set of ($k-n$) \textit{unseen classes}, where $\mathcal{Y} = \mathcal{Y}_{s} \cup \mathcal{Y}_{u},\: \mathcal{Y}_{s} \cap \mathcal{Y}_{u} = \emptyset$. Note that we assign $\mathcal{Y}_{s}$ and $\mathcal{Y}_{u}$ alternately (one every other) on the map. Refer to Figures \ref{fig:zero-shot-gln} for an illustration.

For zero-shot indoor localization, the goal is to enable the system to recognize photos of \textit{unseen classes} $\mathcal{Y}_{u}$ through training only on photos of \textit{seen classes} $\mathcal{Y}_{s}$. It is impossible to employ traditional supervised learning methods to train a model that can recognize the unseen classes without seeing them before. Instead, we leverage the information that is available to both groups (\emph{i.e.}, floor plans) to bridge them together. There are three key steps to perform zero-shot indoor localization: (a) training the Map2Vec location embeddings, (b) learning a compatibility function with the embeddings and GLN, and (c) performing zero-shot recognition.

\subsubsection{Map2Vec Location Embedding.} To overcome the aforementioned problem, we propose the \textit{Map2Vec} mechanism to learn location-aware graph embeddings to correlate the seen and unseen classes. Figure \ref{fig:zero-shot-gln}(a) shows an illustration of this procedure. For a given map (floor plan) with $k$ locations, we define a graph $\mathcal{G'}=(\mathcal{V'}, \mathcal{E'})$ where each vertex $v'$ represents a location and each edge $e'$ is a path between locations. Similar to GLN in the standard setting, we adopt the Graph Neural Networks as in Eq. \ref{eq:gcn} to train graph structure-aware node embeddings and initialize each of the hidden states using the coordinate $o_i \in \mathbb{R}^2$ for location $i$. After $L$ layers of message-passing, we extract the final hidden state $ h_i^L \in \mathbb{R}^k$ as the location embedding for class $y_i$: $\psi(y_i)$.

\subsubsection{Compatibility Function.} To take advantage of the learned location embeddings, we aim at doing \textit{knowledge transfer} so that the indoor localization knowledge can be transferred from seen to unseen classes. To carry out the knowledge transfer, we utilize a compatibility function $F: \mathcal{X} \times \mathcal{Y} \rightarrow \mathbb{R}$ which is a mapping from an image-class pair to a scalar score for the specific class. Figure \ref{fig:zero-shot-gln}(b) shows an illustration of learning the compatibility function. Since only the samples from seen classes are used for learning the compatibility function, it should be in a class-agnostic form. We follow \cite{Sumbul2018} and define the compatibility function in a bilinear form as follows:

\begin{equation}
    F(x,y_{s}) = \phi(\mathbf{x})^\top W \psi(y_{s}),
    \label{compatibility_func}
\end{equation}

\noindent where $\phi(\mathbf{x}) \in \mathbb{R}^d$ is the image representation of an image $\mathbf{x}$ from seen classes, $\psi(y_{s}) \in \mathbb{R}^k$ is the location embedding of a seen class $y_{s}$ and $W \in \mathbb{R}^{d \times k}$ is the weights that we are actually learning. In this context, $W$ is in fact our GLN that takes in $d$-dimensional image representation and output $k$-dimensional logits. Similar to standard indoor localization, we adopt cross entropy loss as the objective function.

\subsubsection{Zero-Shot Recognition.}
Once the compatibility function is learned, we can utilize it to make predictions on unseen classes. Refer to Figure \ref{fig:zero-shot-gln}(c) for an illustration. Zero-shot indoor localization is achieved by assigning the query image a location class $y^*$ that maximizes $F(x, y)$:

\begin{equation}
    y^* = \argmax_{y \in \mathcal{Y}}F(x,y),
\end{equation}

\noindent Unlike \cite{Sumbul2018}, we predict on all possible locations $y \in \mathcal{Y}$ instead of on only unseen classes $\mathcal{Y}_{u}$ to simulate real-world scenarios.

%%%%%%%%%%%%%%%%%%%%%%%%%%% Experiments %%%%%%%%%%%%%%%%%%%%%%%%%%%

\section{Experiments}
In this section, we conduct extensive experiments to evaluate the proposed method. Towards this aim, we first explain the implementation details, evaluation datasets and metrics. We then compare our GLN based indoor localization systems with existing models under both standard and zero-shot settings.

\subsection{Implementation Details}
\label{implementation_details}
We implement our model based on \textit{PyTorch} \cite{paszke2017automatic} framework and train on a single NVIDIA Titan X. To extract image representation, we adopt ResNet-152 \cite{he2016deep} and utilize off-the-shelf weights from {\fontfamily{qcr}\selectfont torchvision} package of PyTorch. We employ data augmentation technique to randomly flip, rotate by 10 degrees and resize to $256 \times 256$ pixels, followed by randomly cropping a patch of $224 \times 224$ pixels. The output of the CNNs is a 2,048-dimensional feature for each image ($d = 2048$). For both of our standard GLN and zero-shot GLN, we adopt Graph Convolutional Networks \cite{kipf2016semi} as the backbone of message passing process and the attention mechanism in Graph Attention Networks (GATs) \cite{velivckovic2017graph}, and we utilize the implementation provided by \textit{PyTorch Geometric} \cite{Fey/Lenssen/2019}. An undirected edge is implemented with two directed edges of opposite directions in the experiments. We observe one layer of message propagation ($L=1$) empirically gives the best performance. The dimension of the latent representation $h^l_i$ is $256$. We utilize ReLU nonlinear activation for the original GLN and adopt LeakyReLU for the attentional GLN at each layer, while both are followed by a batch normalization layer and a dropout layer to stabilize training. Attention mechanism $a$ is implemented with a single FC layer. We train the model in an end-to-end manner with learning rate $3 \times 10^{-4}$ with Adam optimizer of the exponential decay rate 0.9 and 0.999 for the first- and the second-moment estimates respectively. 

\subsection{Evaluation Datasets and Metrics}

\begin{figure}[t!]
\centering
\includegraphics[width=\columnwidth]{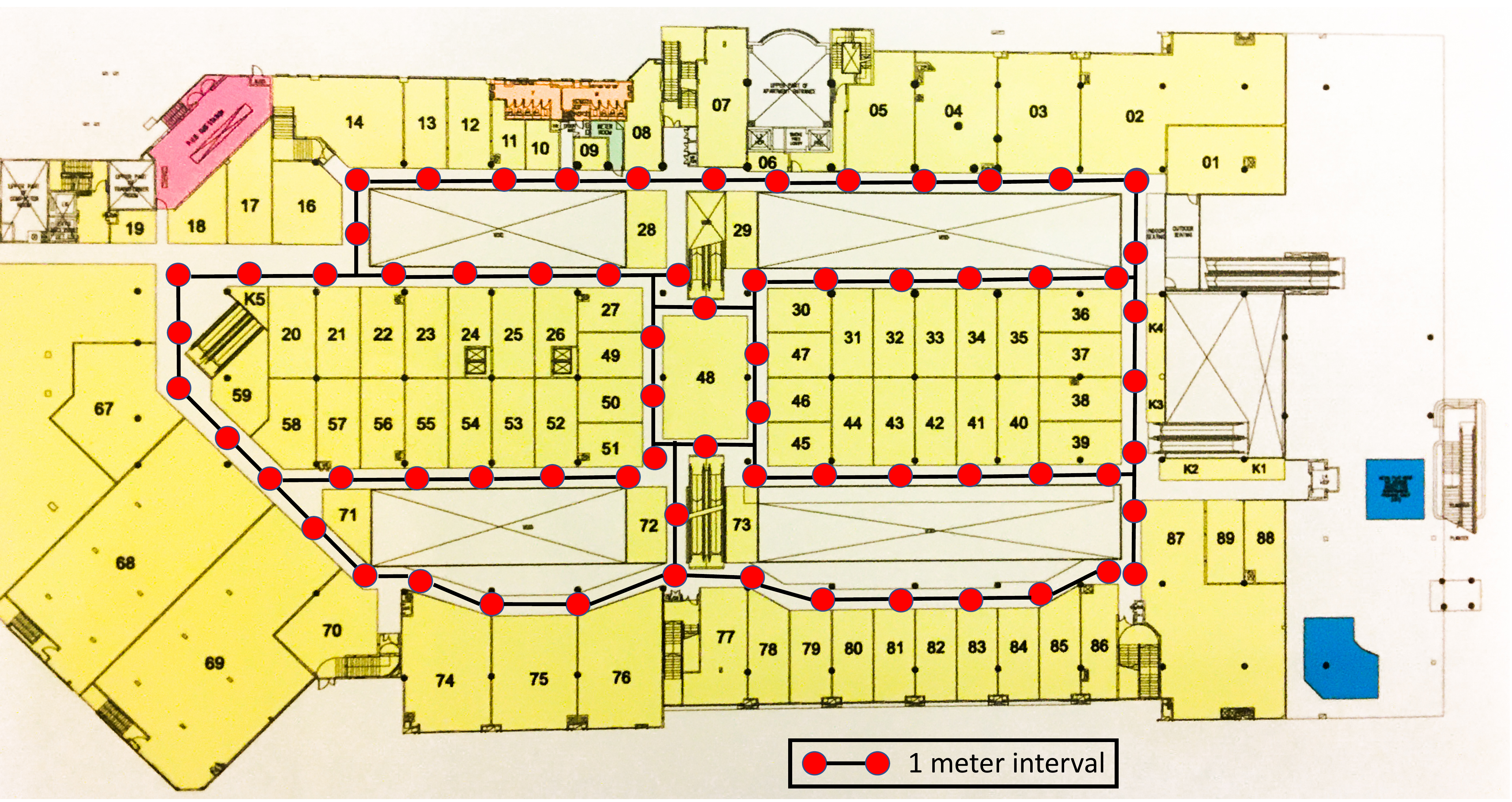}
\caption{An illustration of WCP dataset where the red vertices represent locations, and black edges denote the adjacency of vertices. Note that the locations are not draw to scale and are for illustrative purposes only.}
\label{fig:wcp_map_screenshot}
\end{figure}

We evaluate our proposed method on two datasets: ICUBE \cite{liu2017multiview} that is publicly available and WCP that is collected by ourselves. 

\subsubsection{ICUBE dataset} 
The ICUBE dataset contains 2,896 photos of 214 locations in an academic building. For standard indoor localization, to perform a fair comparison, we closely follow the original paper \cite{liu2017multiview} to divide the dataset into a training set of 1,712 images and test set of 1,184 images. While in the zero-shot setting, we set aside 1,368 images of 102 locations as seen classes, where 1,092 of them are for training and the other 276 of them are for validation during training the compatibility function. The remaining 1,528 images of 112 locations are set as unseen classes to be used in zero-shot recognition.

\subsubsection{WCP dataset} 
The WCP dataset consists of 3,280 photos of 394 locations in a shopping center. We assign 2,624 images for training and the other 656 images for testing in the standard indoor localization experiment. In the zero-shot setting, 1,696 images of 204 locations are assigned as seen classes, in which 1,360 and 336 of them are for training and validation compatibility function, respectively. The other 1,584 images of 190 locations are unseen classes. Overall, WCP is more difficult than ICUBE due to its higher number of classes and more complicated scenes such as shops and restaurants. Both datasets are collected in 1-meter distance interval and have a corresponding map that has vertices of locations and edges of adjacency. Figure \ref{fig:wcp_map_screenshot} shows an illustration of the WCP dataset. % We will release the dataset soon to facilitate researches in efficient image-based indoor localization.

\subsubsection{Evaluation Metrics} 
We report \textbf{one-meter-level accuracy} and \textbf{Cumulative Distribution Function of localization error (CDF@k)} at distance \textit{k}. For zero-shot indoor localization, to perform a more detailed evaluation of the models' strengths and weaknesses, we utilize multiple metrics including \textbf{CDF@k}, \textbf{Recall@k} that sees if the ground truth presents in top \textit{k} predictions ordered by confidence scores, and \textbf{Median Error Distance (MED)} which calculates the error distance of 50-percentile predictions. Note that the distance unit is 1-meter for CDF and MED.

\subsection{Quantitative Results}

\subsubsection{Standard Indoor Localization}

\begin{figure}[t!]
\centering
\includegraphics[width=1.0\columnwidth]{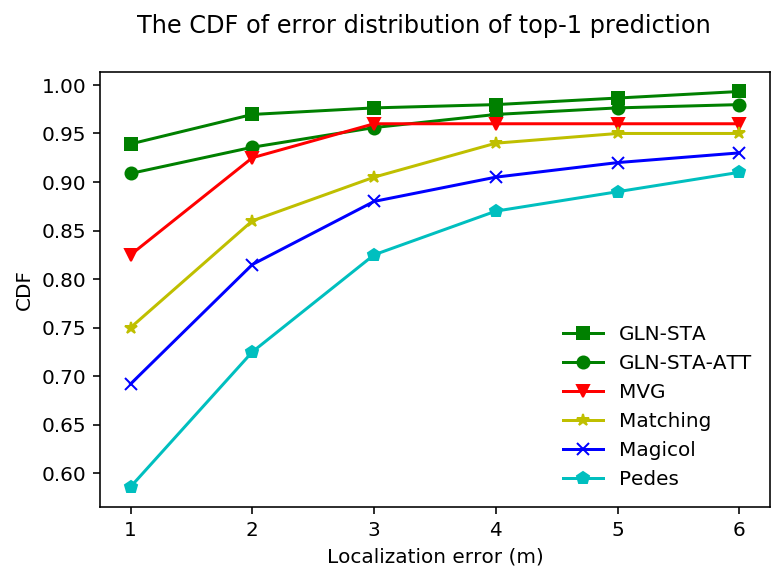}
\caption{The cumulative distribution function (CDF) curves of the localization error of the previous and our approaches in standard indoor localization setting on ICUBE dataset.}
\label{fig:standard-indoor-localization-sota}
\end{figure}

\begin{table}[t!]
    \centering
    \def\arraystretch{1.2}%
    \begin{tabular}{c|l|c}
    \hline
    \centering Dataset & Method & Meter-level Accuracy \\
    \hline
    \centering \multirow{6}{*}{ICUBE} & Pedes \cite{Li:2012:RAI:2370216.2370280} & 58.30\%\\
        & Magicol \cite{shu2015magicol} & 69.20\% \\
        & Matching \cite{Ravi2007} & 75.00\% \\
        & MVG \cite{liu2017multiview} & 82.50\% \\
        & \textbf{GLN-STA} & \textbf{93.92\%} \\
        & \textbf{GLN-STA-ATT} & \textbf{90.88\%} \\
    \hline
    \centering MALL-1\dag & Sextant \cite{Gao2016} & 47\% \\
    \centering MALL-2\ddag & GeoImage \cite{Liang2013} & 53\% \\
    \centering \multirow{2}{*}{WCP} & \textbf{GLN-STA} & \textbf{79.88\%} \\
    &  \textbf{GLN-STA-ATT} & \textbf{79.88\%} \\
    \hline
    \end{tabular}
    \caption{Performance comparison with state-of-the-art models on ICUBE, WCP and the respective MALL datasets. Results of previous approaches on ICUBE are taken from \cite{liu2017multiview}, while results on distinct MALL datasets are taken from their respective papers. \dag MALL-1 consists of 108 locations and 686 images. \ddag Mall-2 contains 20,000 images (locations).}
    \label{table:standard-gln-and-sota}
\end{table}

\begin{table*}[t!]
    \begin{center}
    \def\arraystretch{1.4}%
    \begin{tabular}{c|c|ccccc|ccccc|c}
    \hline
    \centering \multirow{2}{*}{Dataset} & \multirow{2}{*}{Method} & \multicolumn{5}{c|}{Recall@k} & \multicolumn{5}{c|}{CDF@k} & \multirow{2}{*}{MED}\\
    & & k=1 & k=2 & k=3 & k=5 & k=10 & k=1 & k=2 & k=3 & k=5 & k=10 & \\
    \hline
    % \hline
    \centering \multirow{3}{*}{ICUBE} & Baseline-coord & 0.00 & 0.01 & 0.02 & 0.03 & 0.03 & 3.53 & 3.73 & 5.96 & 11.65 & 23.95 & 23.00 \\ \cline{2-13}
    \centering & \textbf{GLN-ZS} & 8.12 & 14.40 & 22.78 & 30.89 & \textbf{46.60} & \textbf{19.90} & 33.77 & \textbf{45.81} & \textbf{56.28} & \textbf{74.87} & \textbf{3.76} \\
    \centering & \textbf{GLN-ZS-ATT} & \textbf{8.38} & \textbf{14.92} & \textbf{23.30} & \textbf{32.20} & 45.81 & 18.59 & \textbf{34.55} & 43.71 & 55.24 & 73.04 & 4.09 \\
    \hline 
    \centering \multirow{3}{*}{WCP} & Baseline-coord & 0.00 & 0.00 & 0.00 & 0.00 & 0.00 & 1.01 & 1.01 & 2.78 & 3.79 & 8.84 & 27.00 \\ \cline{2-13}
    \centering & \textbf{GLN-ZS}& \textbf{2.02} & \textbf{6.06} & 7.83 & 12.37 & \textbf{24.75} & 8.84 & \textbf{13.38} & 17.42 & 22.98 & 50.25 & 9.97 \\
    \centering & \textbf{GLN-ZS-ATT} & \textbf{2.02} & 4.55 & \textbf{8.33} & \textbf{13.64} & 24.50 & \textbf{9.09} & \textbf{13.38} & \textbf{19.70} & \textbf{25.00} & \textbf{51.52} & \textbf{9.93} \\
    \hline
    \end{tabular}
    \caption{Results of zero-shot indoor localization in comparison of \textit{Recall@k}, \textit{CDF@k} and \textit{Median Error Distance} (MED) on ICUBE and WCP datasets. Note that numbers of recall and CDF are in \% (the higher the better), while the numbers of median error distance are in meter (the lower the better). MED results are estimated with linear interpolation.}
    \label{table:zero-shot-result}
    \end{center}
\end{table*}

To compare with existing indoor localization methods, we choose not only those that are purely based on images but also those based on signals. Pedes \cite{Li:2012:RAI:2370216.2370280} is a pedestrian dead reckoning localization method using inertial sensors. Magicol \cite{shu2015magicol} incorporates geomagnetic field and WiFi signal to perform indoor positioning. Matching \cite{Ravi2007} performs image comparison by scoring with multiple off-the-shelf algorithms. MVG \cite{liu2017multiview} is a multi-view localization method via graph retrieval based on images and geomagnetism. Note that Magicol and MVG are not purely image-based methods. GLN-STA is our original GLN and GLN-STA-ATT is the GLN with self attention mechanism. The upper part of Table \ref{table:standard-gln-and-sota} shows the meter-level accuracy compared to existing image-based methods, where our GLN variants surpass the others with significantly higher within one-meter accuracy on the ICUBE dataset and improve the state-of-the-art by 13.8\%.

We observe that the usage of attention mechanism does not help on both datasets. Note that the scale of our quadrilateral graph is very different from the common graph datasets (e.g. citation networks \cite{sen2008collective}, WebKB graphs \cite{webkb}) that have thousands of nodes and edges. Moreover, it was observed in \cite{mostafa2020permutohedral,zhang2018graph} that the common instantiation of the attention mechanism on GNNs (i.e. GATs) does not necessarily bring performance boost over standard GCNs on distinct graph datasets. Therefore, more investigation into the way of instantiating and applying graph attention mechanism in our architecture is needed and we leave it as our future work.

Figure \ref{fig:standard-indoor-localization-sota} shows the full localization error curve (CDF@k), where our GLN perform consistently better than previous approaches, regardless of the requirement of infrastructures.

We also list the additional results of the previous approaches that cannot be reproduced to evaluate on our datasets due to their infrasture requirements.\footnote{Note that since they were evaluated on distinct shopping center datasets, the results may not be directly comparable and serve for reference purposes.} Sextant \cite{Gao2016} leverages image matching algorithms to identify and match with the pre-selected reference objects. GeoImage \cite{Liang2013} performs image matching against a geo-referenced 3D image dataset. Their localization accuracy on the respective shopping mall datasets and our GLN variants on the WCP dataset is showed at the lower part of Table \ref{table:standard-gln-and-sota}. Our GLN-variants achieve significantly better localization performance than the previous methods without any infrastructure requirement.

\subsubsection{Zero-Shot Indoor Localization}

To simulate the real-word case, while we only perform the localization on data of unseen classes $\mathcal{Y}_{te}$, we still make predictions on all possible locations. 
To demonstrate that our proposed approach helps in zero-shot localization, we implement a baseline method Baseline-coord that utilize the coordinates $o$ but not the Map2Vec location embeddings $\psi(y)$ to train the compatibility function. Baseline-coord uses the same standard GLN as the backbone architecture. 

The experimental results of zero-shot indoor localization on the ICUBE and WCP dataset are shown in table \ref{table:zero-shot-result}. GLN-ZS is the original GLN and GLN-ZS-ATT is the attentional GLN, both in the zero-shot setting. On both datasets, GLN-ZS and GLN-ZS-ATT outperform the baseline approach by a large margin. In specific on the ICUBE dataset, GLN-ZS significantly outperforms Baseline-coord by achieving 56.3\% 5-meter accuracy (CDF@5) and median error of 3.76 meters, which are considered promising since all test locations are never seen during training. 

Similar to the observation in the experiments of the standard setting, GLN-ZS-ATT has similar performance to GLN-ZS on both ICUBE and WCP. In addition to the possible reasons we discussed in the previous section, we find that it may also results from the relatively monotonic scenes in the ICUBE dataset so that the attention mechanism does not help much on distinguishing views. In contrast, GLN-ZS-ATT has slight performance improvements over GLN-ZS in terms of CDF@k and MED on WCP.

Overall, compared to experiments on ICUBE, both variants of GLN perform less powerful on the metrics, presumably due to higher variance of scenes and a larger number of classes.

\begin{figure}[t!]
\centering
\includegraphics[width=1.0\columnwidth]{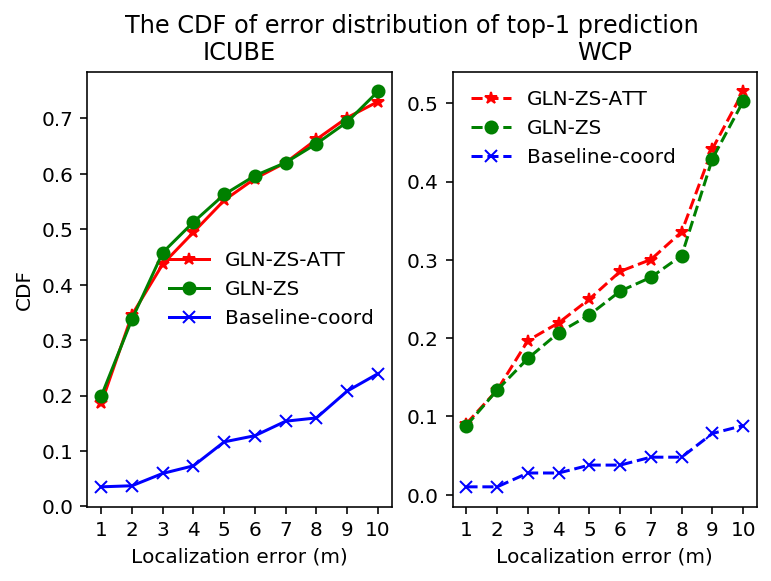} 
\caption{The cumulative distribution function (CDF) curves of the localization error of the zero-shot indoor localization experiments on ICUBE (left) and WCP (right) datasets.}
\label{fig:zero-shot-cdf-result}
\end{figure}

\begin{figure*}[t!]
\centering
\includegraphics[width=1.0\textwidth]{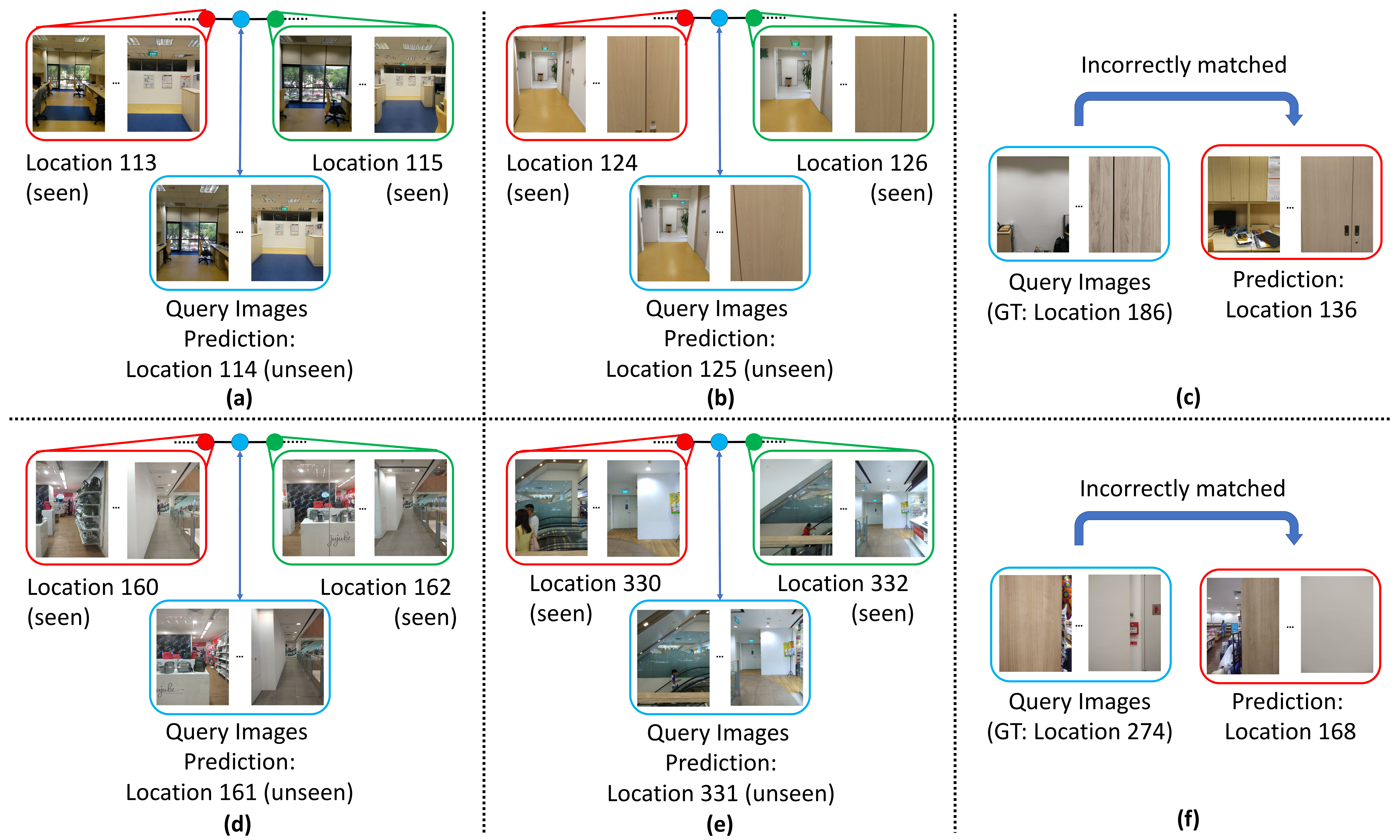} 
\caption{Qualitative results of zero-shot indoor localization on ICUBE (the top row) and WCP (the bottom row) dataset. The first two columns show examples of successful localization cases by utilizing the adjacency of seen classes to unseen classes, where the red, blue and green circles represent three adjacent locations. The last column shows examples of unsuccessful localization cases where our system is misled, especially when there are more query photos lacking distinguishable features.}
\label{fig:zsl_qualitative_results_succs_fail}
\end{figure*}

Figure \ref{fig:zero-shot-cdf-result} shows the full CDF@k curves of zero shot GLN variants and Baseline-coord, where ours perform consistently better. For example on the ICUBE dataset, GLN-ZS shows strong performance improvements ranging from 3.1 to 5.6 times higher CDF@k than the baseline, demonstrating the benefit of the Map2Vec embedding. In addition, as mentioned above, GLN-ZS-ATT is shown to have more consistent performance improvements over GLN-ZS especially on the harder WCP dataset.

\subsection{Qualitative Results for Zero Shot GLN}

To better identify the strengths and weaknesses of our proposed zero-shot approach, we perform qualitative analysis for GLN-ZS in zero-shot indoor localization setting. The left two columns of Figure \ref{fig:zsl_qualitative_results_succs_fail} show examples of successful localization where the correct prediction is made by inferring that the location (\emph{e.g.}, location 114 in Fig. \ref{fig:zsl_qualitative_results_succs_fail}(a)) of the query images is between two neighboring seen locations (\emph{e.g.}, location 113 and 115) by referring to the Map2Vec location embeddings. While some of the views that parallel corridor are extremely similar (\emph{e.g.}, the second view of Fig. \ref{fig:zsl_qualitative_results_succs_fail}(d) and (e)), our model is able to extract robust representation by passing identifiable features from neighboring views (\emph{e.g.}, the first view of Fig. \ref{fig:zsl_qualitative_results_succs_fail}(d) and (e)) to infer the correct location. However, our system could still be misled especially when there are more query photos lacking distinguishable features. The last column (Fig. \ref{fig:zsl_qualitative_results_succs_fail}(c) and (f)) shows unsuccessful localization cases where more images contain no distinguishable features. For instance, in Fig. (c) the first query photo consists of mostly a white wall and the second photo contains merely the surface of a cabinet.

%%%%%%%%%%%%%%%% Conclusion %%%%%%%%%%%%%%%%

\section{Conclusion}
In this paper, we first propose a novel neural network based architecture, namely Graph Location Networks (GLN) to perform multi-view indoor localization. GLN takes in photos of different views and make location predictions based on robust location representations with the message-passing mechanism. To reduce prohibitive labor cost when deployed in large-scale indoor environments, we introduce a novel task named zero-shot indoor localization and propose a effective learning framework which is used to adapt GLN to a dedicated zero-shot version to make predictions on unseen locations. We evaluate our proposed approach not only on the publicly available ICUBE dataset but also on our own benchmark dataset WCP that we make publicly available to facilitate researches in multi-view indoor localization systems. Experimental results show that our proposed method achieves state-of-the-art results in the standard setting and performs well with promising accuracy in the zero-shot setting. 

\begin{acks}
This research is partly supported by the Natural Science Foundation of Zhejiang Province, China (No. LQ19F020001), the National Natural Science Foundation of China (No. 61902348, U1609215, 61976188, 61672460), and Singapore's Ministry of Education (MOE) Academic Research Fund Tier 1, grant number T1 251RES1713.
\end{acks}

%%%%%%%%%%%%%%%% Reference %%%%%%%%%%%%%%%%

\bibliographystyle{ACM-Reference-Format}
\bibliography{zsgln}

\end{document}